%% file: main.tex
\documentclass{article}

\PassOptionsToPackage{numbers, compress}{natbib}

\usepackage[preprint]{neurips_2024}

\usepackage[utf8]{inputenc} 
\usepackage[T1]{fontenc}    
\usepackage{hyperref}       
\usepackage{url}            
\usepackage{booktabs}       
\usepackage{amsfonts}       
\usepackage{nicefrac}       
\usepackage{microtype}      
\usepackage{xcolor}         
\usepackage{amsmath}
\usepackage{graphicx}
\usepackage{placeins}

\title{Challenges in Deploying Long-Context Transformers: A Theoretical Peak Performance Analysis}

\author{%
  Yao Fu \\
  University of Edinburgh\\
  \texttt{yao.fu@.ed.ac.uk}
}

\begin{document}

\maketitle

\begin{abstract}
  Transformer-based long context  generative models power emerging AI applications like 
  hour-long video understanding and project-level coding agent. 
  Deploying long context transformers (e.g., 100K to 10M tokens) is prohibitively expensive compared to short context (e.g., 4K tokens) model variants. 
  Reducing the cost of long-context transformers is becoming a pressing research and engineering challenge starting from the year of 2024.
  This work describes a concurrent programming framework for quantitatively analyzing the efficiency challenges in serving multiple long-context requests under limited size of GPU high-bandwidth memory (HBM) regime. 
  We give a detailed analysis of how all additional computational costs, compared to 4K context, trace back to \textit{one single source: the large size of the KV cache}. 
  We use a 34B GPT-3.5 level model of 50K context on A100 NVLink as a running example, and describe how its large KV cache
  causes four types of deployment challenges: 
  (1) prefilling long inputs takes much longer compute time and GPU memory than short inputs;
  (2) after prefilling, the large KV cache residing on the GPU HBM substantially restricts the number of concurrent users being served; 
  (3) during decoding, repeatedly reading the KV cache from HBM to SM largely increases latency; 
  (4) when KV cache memory overflows, swapping it from HBM to DDR causes significant context switching latency.
  We use this framework to analyze existing works
  and identify possibilities of combining them to build end-to-end systems.
  Overall, this work offers a foundational framework for analyzing long context transformer deployment and identifies directions towards reducing the inference cost of 1M context to be as cheap as 4K. 
\end{abstract}

\section{Introduction}
\label{sec:intro}
\input{000_intro}

\section{Challenges in Deploying Long-Context Transformers}
\label{sec:challenges}

\input{010_challenges}
\section{Compressibility Analysis and Directions of Optimization}
\label{sec:compressibility}
\input{020_compressibility}


\section{Conclusions}
\label{sec:conclusion}
\input{040_conclusion}

\bibliographystyle{plainnat}
\bibliography{deploying_long_context}




\end{document}

%% file: 000_intro.tex
Suppose one has successfully turned an open-source GPT-3.5 to GPT-4 level language models (e.g., LLaMA 3~\citep{touvron2023llama}, DeepSeek~\citep{DeepSeekAI2024DeepSeekV2AS}, QWen~\citep{bai2023qwen} and Mixtral~\citep{jiang2024mixtral}) into a long-context variant (e.g., through continual pretraining~\citep{fu2024data} and instruction tuning~\citep{bai2024longalign}) then wants to deploy this model for a wide spectrum of applications like repository-level coding and hour-long video understanding. 
These applications typically require the input-context to be of 100K to 10M tokens where the paradigm changes substantially from the 4K short-context regime. 
We are interested in an ambitious goal: 
\begin{center}
    \textit{How to reduce the deployment of 1M context production-level transformers to be as cheap as 4K?}
\end{center}
Why serving long-context transformers is expensive? Most of the cost overhead trace back to \textit{one single source: the size of the KV cache}. 
Consider a 30+B 100K context GPT-3.5 quality open-source models like QWen or Yi, the differences between KV cache for 4K v.s. 100K context is:
\begin{align}
    \text{100K context:}\quad\quad &\underset{\texttt{seqlen}}{100000} \times \underset{\texttt{layer}}{60} \times \underset{\texttt{head}}{8} \times \underset{\texttt{dim}}{128} \times \underset{\texttt{KV}}{2}\times \underset{\texttt{bf16}}{2}\;\text{bytes} = 22.8 \texttt{GB}\\
    \text{4K context:}\quad\quad &\underset{\texttt{seqlen}}{4000} \times \underset{\texttt{layer}}{60} \times \underset{\texttt{head}}{8} \times \underset{\texttt{dim}}{128} \times \underset{\texttt{KV}}{2}\times \underset{\texttt{bf16}}{2}\;\text{bytes} = 0.91 \texttt{GB}
\end{align}
Here we use the Yi-34B 200K~\citep{young2024yi} configuration (60 layers, 8 kv heads and 128 hidden dimension). Suppose we use 2 $\times$ 80G A100 tensor parallelism to serve this model in bf16, then we have $2\times80 - 34\times2 = 122$ GB spare space for storing the KV cache. 
From this first glance, we immediately realize that under this setting, we can achieve about 100+ users concurrency of 4K context, but only 5 users of 100K context.
If we were able to deploy 1M context models as cheap as 4K, we can substantially democratize long-context and multimodal generative models and foster an emerging application ecosystem. 

\input{figures/fig_programming_model}

This work gives a full-stack quantitative analysis of challenges in deploying long-context transformers, from hardware to system, from model architecture to user preference.
We first describe a concurrent programming framework (Fig~\ref{fig:programming_model}) for serving multiple long-context user requests under limited GPU HBM size. 
We define session-based throughput, the rounds of user interactions in a given period, as the end-to-end evaluation, and decompose it into four key metrics: the level of concurrency, the prefilling latency, the decoding latency, and the context switching overhead. 
We discuss how these four metrics are bounded by the size of HBM, the GPU flops, the HBM bandwidth, and the PCIE bandwidth respectively, and how these challenges eventually trace back to the size of the KV cache, leading to the core research problem about lossless compression of the KV cache. 
We further discuss how hardware architectures change the performance factors, 
how existing works compress the KV cache from the layer, head, token, and hidden state dimension, 
and how the relative cost between prefilling and decoding changes. 
We hope this work serves as the fundamental framework towards reducing the inference cost of 1M context to be as cheap as 4K.

%% file: figures/fig_programming_model.tex
\begin{figure}[t!]
    \centering
    \includegraphics[width=\linewidth]{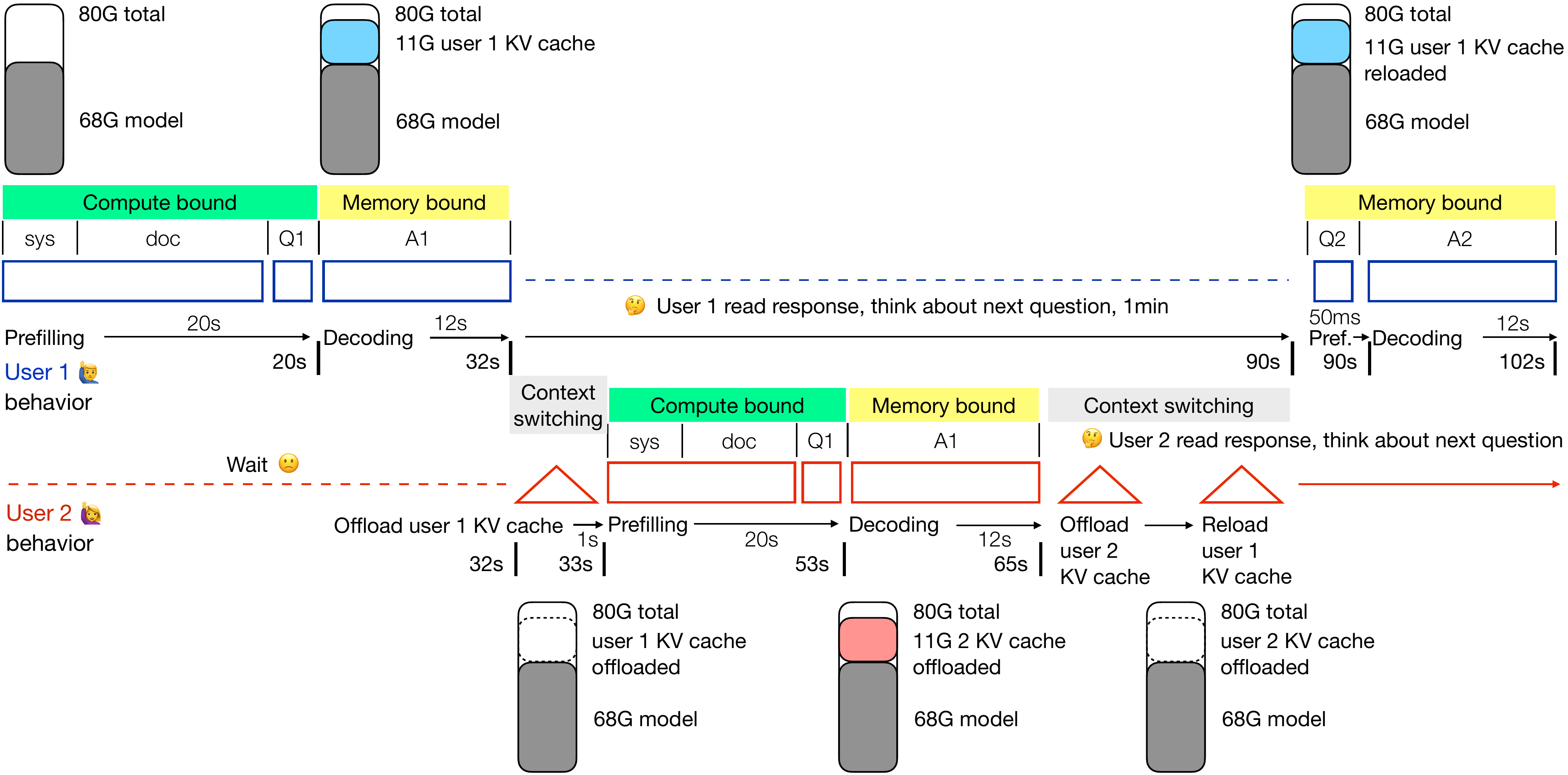}
    \caption{A concurrent programming framework for serving multiple long-context user requests under limited GPU HBM size. 
    There are four key factors collectively determine the overall throughput of user interaction sessions: 
    (1) \textit{concurrency is bounded by the HBM size}: the number of concurrent user being served is bounded by the size of the GPU high-bandwidth memory (HBM);
    (2) \textit{prefilling is compute bound}: the latency to the first generated token, i.e., the prompt prefilling time, is bounded by the floating point operation per second (flops) of the GPU; 
    (3) \textit{decoding is memory bound} (under critical batch size): the latency (generated token per second) of autoregressive decoding is bounded by the bandwidth of the HBM; 
    (4) \textit{context switching is PCIE bound}: offloading user 1's KV cache to the CPU DDR and loading user 2's KV cache to the HBM is bounded by the PCIE bandwidth. 
    Compared to the short-context setting, concurrency and context-switching challenges are much more severe under a long-context setting. 
    All efficiency challenges from these four key factors eventually trace back to the size of the KV cache.
    }
    \label{fig:programming_model}
\end{figure}

%% file: 010_challenges.tex
Given a production-level long-context transformer, our objective is to reduce the deployment cost of 1 million token context to be as cheap as 4K. 
We consider 30+B models of 50K context and group query attention (GQA) on 80G A100 NVLink as a running example because they are typically of GPT-3.5 capability, specifically we use Yi-34B 200K configuration because, by the time of writing this paper, it is the only open-source 30+B model supporting long-context. 
We first describe a concurrent programming framework about the workflow for serving multiple long-context user queries (Fig.~\ref{fig:programming_model}) and identify four key performance metrics: 
concurrency, prefilling, decoding, and context-switching.

We focus on \textit{theoretical peak performance} analysis. 
This is to say, we assume we already have an efficient implementation that can squeeze out all hardware performance
(which typically requires sophisticated engineering efforts like cuda kernel programming, memory management, model parallelism .etc), 
and study under an efficient enough implementation, what are the key challenges and limits we should tackle. 
Although we do not have an actual implementation (whose engineering effort is quite nontrivial, e.g., see vLLM~\citep{kwon2023efficient} and InfiniteLLM~\citep{lin2024infinite}, thus significantly beyond the scope of our current resource),
theoretical peak performance (which is widely used in analyzing LLM serving systems) is an important tool to identify system bottlenecks and provides guidance to further advance efficiency (e.g., most existing serving systems focus on decoding yet we may want to focus more on prefilling when the context is longer than 200K, as we discuss later in Fig.~\ref{fig:prefill_dec}).

\subsection{A Concurrent Programming Framework under Limited GPU HBM Size}
\label{ssec:programming_model}

\input{tables/tab_user_behavior}
\textbf{Concurrent User Interaction Sessions and Preferences}\quad\quad 
As is shown in Table~\ref{tab:user_behavior} and Fig.~\ref{fig:programming_model}, in a typical interaction session, a user starts from a prompt of a long document followed by a question, and feeds it to the model. 
The model receives the initial prompt, \textit{prefill} it to become the KV cache. 
The user waits for the prefilling stage until the first token starts to generate and prefers the waiting time not to be so long. 
After prefilling, the model starts autoregressive \textit{decoding}.
The user reads the output simultaneously with the decoding process and prefers the decoding to be faster than the human reading speed. 
After the model finishes decoding, the user continues to read the response, think, maybe take a sip of coffee, and then start to type the next question.
The follow-up prompts are usually not as long as the first prompt because the first prompt typically contains the long context (book or video) while the follow-up prompts are typically question-only.
When the first user is reading the model response and thinking about what to ask next, the model is essentially idle, so at the same time if another user asks another question, the model could do \textit{context switching} by offloading the first user's KV cache to the CPU DDR memory to make HBM space to store the second user's KV cache. 
The two users ask follow-up questions interactively until their session ends.

\textbf{Session-Based Throughput as an End-to-End Objective}\quad\quad
We consider the situation where multiple users simultaneously interact with the model. 
Assume on average, a user session consists of the document/ video of 50K tokens and 5 rounds of questions.
After receiving the answer to the previous question, the user spends about 1 minute reading the answer and thinking about the next question. 
Our objective is to maximize a session-based throughput defined as:
\begin{align}
    \texttt{throughput} = \frac{\texttt{number of user interaction sessions}}{\texttt{time}}
\end{align}
Note that our \textit{session-based} throughput objective is different from existing \textit{token-based} throughput~\citep{kwon2023efficient, li2024snapkv, sun2024triforce} (i.e., number of prefilled or decoded tokens in a given period). 
As we will see soon, token-based throughput is only part of the problem.
Our session-based throughput, i.e., the number of concurrent user interactions in a given period, is an \textit{end-to-end} objective, because we not only consider prefilling and decoding, but also consider memory restrictions and context switching which significantly influence concurrency.

\textbf{Compute v.s. Memory Boundedness, Arithmetic Intensity and Critical Batch Size}\quad\quad
One important observation of transformer inference is that prefilling is usually bounded by the GPU compute power, i.e., the flops, while decoding is bounded by the HBM memory bandwidth. 
We say an operator is compute bound if most of the time of finishing this operator is computing it on GPU's streaming multiprocessor (SMs, where GPU performs block-wise parallel computation).
We say an operator is memory bound if most of the time for finishing this operator is moving the data from the memory to the SMs (instead of actually computing it on the SMs). 
Whether an operator is compute or memory bound depends on its \textit{arithmetic intensity}, defined as how
many floating point operation (FLOP) is performed per memory access operation (IO):
\begin{align}
    \texttt{arithmetic intensity} = \frac{\texttt{FLOP}}{\texttt{IO}}
\end{align}
The higher level of parallelism, the higher flop per memory access, the more likely an operator is compute bound, the better we utilize the hardware. 
On a given GPU, the critical arithmetic intensity, i.e., the level of parallelism, for an operator to change from memory to compute bound is the ratio of its flop / memory bandwidth. For A100 it is:
\begin{align}
    \texttt{A100 critical arithmetic intensity} = \underset{\texttt{A100 bf16 FLOP}}{312\text{T flop per sec}}\;/ \underset{\texttt{A100 HBM Bandwidth}}{2\text{T byte per sec}} = 156
\end{align}
For transformers, the level of parallelism is approximately how many tokens we feed into it, i.e., the batch size. 
This is to say, for A100, when our batch size is larger than 156 tokens, e.g., during prefilling, the user's prompt is 50K tokens, we are compute bound and fully utilizing A100's compute power. 
When our batch size is smaller than 156 tokens, e.g., during decoding we only decode one token at a forward pass, we are memory bound and not fully utilizing A100's compute power.

\textbf{Prefilling}\quad\quad
Now we analyze how long prefilling on A100 takes exactly. 
Since prefilling is compute bound, i.e., context length longer than 156 on A100, its theoretical peak latency is 
\begin{align}
    \texttt{theoretical peak latency} = \frac{\texttt{FLOP of prefilling}}{\texttt{FLOP per second of A100}} \quad\quad \text{If compute bound.}
\end{align}
For a prompt of 50K context length it is (see~\citet{kaplan2020scaling} for a detailed derivation)
\begin{align}
\underset{\texttt{prefilling FLOP}}{4.33\text{P Flop}} &= 
\underset{\texttt{seqlen}}{50 \text{K}} \times (2 \times \underset{\texttt{model param}}{34\text{B}} + 2 \times \underset{\texttt{layer}}{60} \times \underset{\texttt{seqlen}}{50\text{K}} \times\underset{\texttt{hidden}}{4096})\\
\underset{\texttt{prefilling latency}}{14.1 \;\text{seconds}} &= \underset{\texttt{prefilling flop}}{4.33\text{P Flop}} /\; \underset{\texttt{A100 bf16}}{312\text{T Flop per sec}} 
\end{align}
Since 14.1 seconds is the theoretical peak, in Fig.~\ref{fig:programming_model} we round it to 20 seconds to account for the implementation overhead. 
This means the actual implementation may achieve 14.1 / 20 $\approx$ 70\% of the theoretical peak performance, which is a common experience for cuda programming on A100.

If the context length is 4K instead of 50K, then repeating the above computation we get the latency 0.89 seconds. The difference here is
\begin{align}
    \texttt{prefilling latency for 4K} &= 0.89\;\text{seconds}\\
    \texttt{prefilling latency for 50K} &= 14.1\;\text{seconds}\label{eq:prefilling}
\end{align}
The 13-second gap, rooted from the additional flop from the long context, is what we eventually want to close.

\textbf{Decoding}\quad\quad
Now we analyze how long decoding takes exactly.
Since decoding is memory bound, i.e., batch size less than 156 on A100, the theoretical peak latency is
\begin{align}
    \texttt{theoretical peak latency} = \frac{\texttt{bytes of memory access}}{\texttt{A100 HBM bandwidth}} \quad\quad \text{If memory bound.}
\end{align}
For decoding, one forward pass means
\begin{align}
    \texttt{bytes of memory access} = \texttt{model weights} + \texttt{KV Cache}
\end{align}
We assume on average the model generates one screen tokens (typically the user prefers the generation length right about one screen), i.e., about 250 tokens, then the peak latency is
\begin{align}
    \underset{\texttt{one screen tokens}}{250} \times \underset{\texttt{model weight}}{(68\text{GB}} + \underset{\texttt{50K ctx KV cache}}{11\text{GB})}\;/\;\underset{\texttt{A100 HBM bandwidth}}{2\text{T Bytes per sec}} = \underset{\texttt{decoding latency}}{9.8\;\text{seconds}} \label{eq:decoding}
\end{align}
Since 9.8 seconds is theoretical peak, in Fig.~\ref{fig:programming_model} we round it to 12 seconds to account for the implementation overhead. 
If the sequence length is 4K, then its corresponding KV cache is only 0.91GB and the decoding latency reduces to 8.5 seconds. 
Yet if the sequence length increases to 200K, the KV cache becomes 44GB, the latency increases to 14 seconds.
The relative latency increase is correlated with the relative size between the KV cache and the model size, and we eventually want to close it.

\textbf{Concurrency Control and Context Switching}\quad\quad
Another important consideration is that when the KV cache becomes large, the number of concurrent users' cache that the GPU HBM can hold is 
\begin{align}
    \texttt{level of concurrency} = \frac{\texttt{HBM size} - \texttt{model weights}}{\texttt{KV cache}}
\end{align}
This means that concurrency is bounded by the size of the HBM.
Continuing with our 34B 50K model example, if we deploy it on one 80GB A100 we can only serve one user (Fig.~\ref{fig:programming_model}). 
But if the context is 4K, the KV cache is only about 1GB, and we can concurrently serve about 20 users. 

When the second user comes to ask a question about a long document, 
to make room for their KV cache, we need to do context switching: offloading the first user's KV cache to the CPU memory, and loading the second user's KV cache (Fig.~\ref{fig:programming_model}).
This induces the context switching overhead:
\begin{align}
    \texttt{context switching overhead} = \frac{\texttt{user 1's KV cache + user 2's KV cache}}{\texttt{PCIE bandwidth}}
\end{align}
This is to say, the context switching overhead is bounded by the PCIE bandwidth, i.e., how fast the GPU HBM is connected to the CPU DDR. Suppose we use PCIE gen 4 of 20G bytes per second, then the context switching overhead of 2 users of 50K context is:
\begin{align}
    \underset{\texttt{context switching for 50K}}{1.1\;\text{seconds}} = \underset{\texttt{user 1's KV cache}}{(11\text{G bytes}} + \underset{\texttt{user 2's KV cache}}{11\text{G bytes})} / \;\underset{\texttt{PCIE bandwidth}}{{20\text{G bytes per sec}}}
\end{align}
In Fig.~\ref{fig:programming_model} we round the 1.1 seconds to 2 seconds to account for the engineering overhead.
As mentioned earlier, in our setting we can serve 20 users of 4K context length \textit{without context switching} because the HBM is enough to hold their KV cache. 
If we like to increase our 50K concurrency to 20, then the overall context switching overhead also increases with concurrency: 
\begin{align}
    \underset{\texttt{Context switching for 20 users for 50K}}{22\;\text{seconds}} = \underset{\texttt{concurrency}}{20 \;\texttt{users}} \times \underset{\texttt{context switching between 2 users}}{1.1\;\text{seconds}} \label{eq:context_switching}
\end{align}
Note that this 22 seconds overhead does not exist in the 4K context regime.

\textbf{Summary so far}\quad\quad
we have discussed most of the details when deploying long-context transformers using the 34B model 50K context as the running example. 
We see that the overall performance, measured by user interaction throughput, decomposes to four metrics: 
(1) prefilling latency bounded by the GPU flops; 
(2) decoding latency bounded by the HBM bandwidth;
(3) level of concurrency bounded by the size of the HBM; 
(4) context switching overhead bounded by the GPU-CPU connection bandwidth, i.e., the PCIE. 
In the next sections, we will discuss how these metrics change with common factors, and identify the bottleneck eventually trace back to the size of the KV cache.

\input{figures/fig_perf_factors}
\subsection{Factors that Strongly Influence the Performance Metrics}
We start from two basic factors: context length and hardware architecture. 
When increased from 4K to 50K, we show the four metrics (prefilling, decoding, concurrency and context switching) changes with different rate (linear, inverse, and quadratic).
We further show that tensor parallelism improves concurrency, prefilling and decoding, but does not improve context switching. 
Sparse upcycling to a larger mixture of experts reduces concurrency, prefilling and decoding, but does not change context switching. 
The type of attention, i.e., multi-head, multi-query or group query attention, influence performance significantly because they directly influence the size of the KV cache.

\textbf{Context Length}\quad\quad
In the first row of Fig.~\ref{fig:perf_factors}, we compute the theoretical peak performance of the four metrics w.r.t the context length for our Yi 34B running example using the equations in Sec.~\ref{ssec:programming_model}. 
We observe:
(1) concurrency \textit{inversely decreases} with longer context length;
(2) prefilling latency \textit{quadratically increases} with longer context length. 
In comparison, decoding latency and context switching overhead only \textit{linearly increases} with longer context length, and the decoding latency is the least influenced factor because 50K context KV cache is still relatively smaller than model parameters (11GB v.s. 68GB). 
Concurrency and prefilling are the two most severely influenced metrics.

\textbf{Hardware Architecture}\quad\quad
Can we improve the performance by simply using more advanced hardware? In Fig.~\ref{fig:perf_factors} second row, we show the performance improvement tendency with hardware advancements. 
We observe: 
(1) concurrency \textit{linearly increases} with the size of the HBM;
(2) prefilling latency \textit{inversely reduces} with the increased flops when upgrading the device from 4090 to A100 to H100; 
(3) decoding latency \textit{inversely reduces} with the increased memory bandwidth; 
(4) context switching overhead \textit{inversely reduces} with the increased PCIE bandwidth. 
Note that numbers are based on the newest advances by May 2024, and even if we use the newest hardware, the cost gap between 50K and 4K are not closing. 
This is to say, \textit{we cannot count on hardware advances for reducing the cost of serving long-context models}, and we have to make algorithmic innovations.

\textbf{Tensor Parallelism}\quad\quad
utilizes multiple devices for accelerating inference with negligible communication overhead. 
Linearly increasing the number of devices to 2, 4, and 8 introduces more HBM space, thus linearly increasing concurrency. 
Since we equally divide the computation on multiple devices, the prefilling and decoding latency also reduces inversely with the number of GPUs. 
However, \textit{tensor parallelism cannot reduce the context switching overhead} because the PCIE bandwidth between the DDR to the HBM is shared by all devices.

\textbf{Upcycling to Mixture of Experts}\quad\quad
suppose we upcycle our 34B model to be 8$\times$34B mixture of experts model, how would the metrics change? 
The first observation is the model size: since MoE models are much larger than dense models, they will take up more HBM spaces thus reducing concurrency. 
The prefilling and decoding latency changes with the number of activated experts, which is usually 2, thus these two latency will increase approximately by 2 times. 
Since MoE does not change attention, the size of the KV cache does not change, meaning that the context switching overhead does not change. 
In summary, upcycling to MoE changes concurrency, prefilling and decoding latency, but does not change context switching.

\textbf{Types of Attention}\quad\quad
The last but probably so far the most important factor is the type of attention.
Specifically, for the 34B Yi model we consider, it uses group-query attention (GQA) with 8 kv heads but 32 query heads. 
If we increase its kv heads to 32 (i.e., the original multi-head attention, MHA), its corresponding 50K tokens KV cache are: 
\begin{align}
    \text{8 kv heads (GQA)}\quad\quad &\underset{\texttt{seqlen}}{50000} \times \underset{\texttt{layer}}{60} \times \underset{\texttt{head}}{8} \times \underset{\texttt{dim}}{128} \times \underset{\texttt{KV}}{2}\times \underset{\texttt{bf16}}{2}\;\text{bytes} = 11.4 \texttt{GB}\\
    \text{32 kv heads (MHA)}\quad\quad &\underset{\texttt{seqlen}}{50000} \times \underset{\texttt{layer}}{60} \times \underset{\texttt{head}}{4} \times \underset{\texttt{dim}}{128} \times \underset{\texttt{KV}}{2}\times \underset{\texttt{bf16}}{2}\;\text{bytes} = 45.6 \texttt{GB}
\end{align}
In practice people do not often use multi-query attention (MQA) due to its suboptimal performance. 
But compared to the original MHA, GQA directly gives 4x KV cache reduction, translating to 4x improvements of concurrency and 4x context-switching. 
As for decoding latency, we have: 
\begin{align}
    \text{GQA Decoding:}\quad\quad\underset{\texttt{one screen tokens}}{250} \times \underset{\texttt{model weight}}{(68\text{GB}} + \underset{\texttt{GQA KV}}{11\text{GB})}\;/\;\underset{\texttt{A100 HBM bandwidth}}{2\text{T Bytes per sec}} = \underset{\texttt{decoding latency}}{9.8\;\text{seconds}}\\ 
    \text{MHA Decoding:}\quad\quad\underset{\texttt{one screen tokens}}{250} \times \underset{\texttt{model weight}}{(68\text{GB}} + \underset{\texttt{MHA KV}}{46\text{GB})}\;/\;\underset{\texttt{A100 HBM bandwidth}}{2\text{T Bytes per sec}} = \underset{\texttt{decoding latency}}{14.3\;\text{seconds}}
\end{align}
which is about 1.5x decoding latency improvements. 
Combining the improvements in concurrency, decoding and context switching, it is clear that GQA significantly improves long-context efficiency.

\subsection{Most Challenges Trace Back to the Size of the KV Cache} 
We have the following important observations when comparing 50K to 4K in Sec.~\ref{ssec:programming_model}:
(1) to prefill the long input and produce the KV cache, the prefilling latency increases from 0.89 to 14.1 seconds (Eq.~\ref{eq:prefilling});
(2) because the large KV cache residing on the GPU memory, the concurrency reduces from about 20 to 1;
(3) during decoding, repeated loading the KV cache increases the latency from 8.5 to 9.8 seconds (Eq.~\ref{eq:decoding});
(4) the large KV cache induces expensive context-switching overhead, for 20 users it takes about additional 22 seconds (Eq.~\ref{eq:context_switching}). 
These four factors collectively induces significant cost in terms of the end-to-end session-based throughput. 
Our eventual goal is to make the deployment of 1M context as cheap as  4K, and 4K tokens are about 1GB KV cache. 
Then our observations point to one key research problem: 
\begin{center}
    \textit{How to efficiently compress the 1M token KV cache to 1G bytes in a lossless way?}
\end{center}

%% file: tables/tab_user_behavior.tex
\begin{table}[t!]
    \caption{\label{tab:user_behavior}A typical user-model interaction session and important preference constraints. Our setting considers concurrently serving multiple users of similar behaviors and maximize the throughput. 
    }
    \begin{tabular}{ll}
    \toprule
      \bf Action & \bf Preference \\ 
      1. Input a long document and the first question. & - \\
      2. Wait for the model to prefill the prompt. & Not wait for too long. \\ 
      3. The model generates a response. & Generation should be faster than reading. \\ 
      4. Read the response and think about next question. & - \\ 
      5. Ask the next question. & - \\ 
      6. Wait for the model to generate the second response. & Not wait for too long.\\ 
      7. The model generates the next response. & Generation should be faster than reading. \\
      8. Repeat & - \\ 
      \bottomrule
    \end{tabular}
\end{table}

%% file: figures/fig_perf_factors.tex
\begin{figure}[t!]
    \centering
    \includegraphics[width=\linewidth]{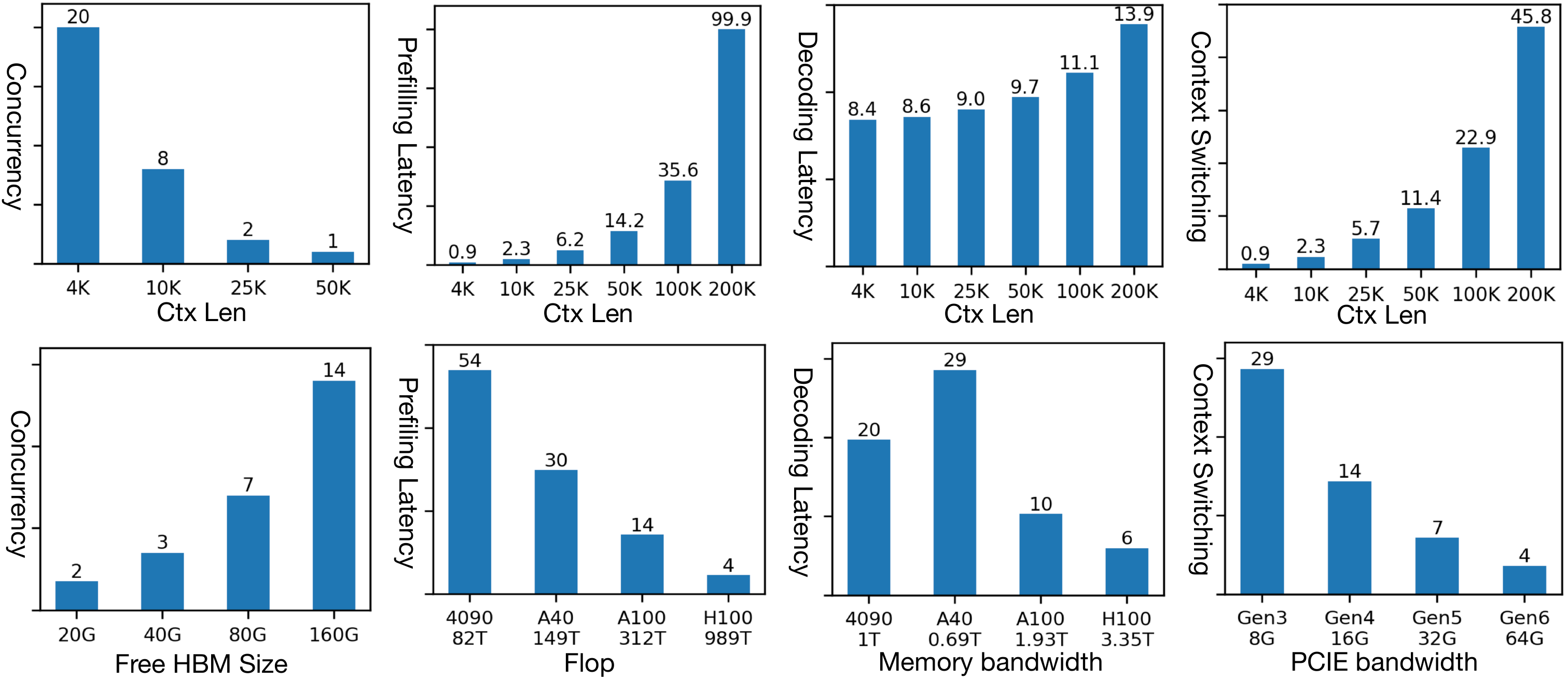}
    \caption{First row: how context length changes the four key performance metrics. 
    Increasing the context length from 4K to 50K inversely reduces concurrency, quadratically increases prefilling latency, linearly increases context switching overhead, and slightly (but also linearly) increases decoding latency. 
    Second row: how different generations of hardware influence the metrics. 
    }
    \label{fig:perf_factors}
\end{figure}

%% file: 020_compressibility.tex
In this section we discuss the potential dimensions to compress the KV cache. 
We first note that without any compression, storing 1M token into bytes only takes about 3 - 10MB disk storage
(depending on the size of the tokenizer's vocabulary), 
so 1GB is more than enough to store the full information of the input tokens. 
The problem is how to make their compressed representations usable by large transformers. 
We start from analysizing the compressibility from layer, head, token, and hidden dimensions, then we discuss the relative cost between prefilling and decoding.

\subsection{Lossless Compressibility by Layer, Head, Token and Hidden}

\input{tables/tab_existing_work}
We first discuss the notion of ``lossless''.
Practitioners usually test the model on a variety of long-context tasks to check if the compression is lossy, 
amoung which the Needle-in-a-Haystack test, which asks the model to precisely retrieve the given information put at arbitrary location of a long context, serves as an \textit{entry barrier}: 
if a model cannot pass this test, we do not trust it can do harder tasks. 
Unfortunately, it seems that two important model families, state-space models (e.g., Mamba~\citep{gu2023mamba}) and linear attention (e.g., LongT5~\citep{guo2021longt5}), cannot pass the needle test, 
so we do not include them into our discussion. 
Recent work from~\citet{wu2024retrieval} shows that there exists a set of special attention heads responsible for retrieving imporant information from the context. 
Their discovery indicates that at least for some layer and some head, the full attention over most of the input tokens should be retained. 
Below we discuss the compressibility of the KV cache from its four dimensions: layer, head, token and hidden.
Table~\ref{tab:existing_work_kv_compression} lists some great existing efforts and what metrics they improve. 

\textbf{Layer}\quad\quad Here the basic hypothesis is that some tasks may not require the full-depth computation~\citep{elhoushi2024layer}.
Skipping some layers during prefilling could be beneficial to all four metrics because it simultaneously reduces prefilling flops and the size of the KV cache. 
In fact, the layer dimension may be radically reduced from the results of existing works like~\citet{wu2022memorizing, Sun2024YouOC}, and it might be possible to \textit{only keep one layer} KV cache for long-context tasks, which is a 1/60 compression ratio. 

\textbf{Head}\quad\quad Here the basic hypothesis is that some heads are specialized for retrieval and long-context related capabilities~\citep{wu2024retrieval}, so it may be possible to retain retrieval heads and prune others. 
Note that head pruning typically happens \textit{after} prefilling, meaning that they only improve decoding, concurrency and context-switching, but prefilling remains the same expensive. 
In general, it seems that at the head dimension is of very high level sparsity, and the number of heads might be possible to radically removed, e.g.,~\citet{wu2022memorizing} show the number of strongest retrieval heads is less than 20, which could potentially translate to a 20 / 1024 compression ratio. 

\textbf{Token}\quad\quad For the token dimension, the basic hypothesis is that if information of a token can be inferred from its context, we can compress this token by either dropping it~\citep{zhang2024h2o} or merging it with its neighbors~\citep{nawrot2024dynamic}. 
Most of the compression at the token dimension does not much improve prefilling, but they typically improves concurrency, decoding and context switching. 
Currently, it seems that the token dimension might not be as sparse as the layer and head dimension because most of the tokens have to be retained for precise retrieval. 
We have not yet seen any work showing the potential of more than 50\% compression ratio on the token dimension. 

\textbf{Hidden}\quad\quad There is not much work on reducing hidden dimension except for quantization~\citep{liu2024kivi, yue2024wkvquant}, presumably because the hidden size is already 128, too small to be reduce further. 
Yet it may still be worth trying applying dimension reduction like LoRA~\citep{hu2021lora} on the KV cache, particularly given recent progress from DeepSeek V2~\citep{DeepSeekAI2024DeepSeekV2AS} which introduces LoRA-like idea that effectively reduces KV head.

An important caveat here is that many existing works may \textit{only emphasize one aspect of the problem}.
For example, TriForce~\citep{sun2024triforce} only considers the decoding latency using speculative decoding.
It does not make the KV cache smaller and even has a tradeoff of increased GPU memory consumption from the additional KV cache from the draft model.
Many existing works are also orthogonal, such that their advantages from different aspects may join force. 
For example, if we could reduce the KV cache to be only 1 layer or 10 heads and keep only 50\% of the tokens, we will have about 1000x performance improvements.
This naturall leads to one call for research: 
\begin{center}
    \textit{Can we integrate existing efforts into an end-to-end system and push full-stack optimization?}
\end{center}

\input{figures/fig_prefill_dec}
\subsection{Relative Cost Between Prefilling and Decoding}
To end-to-end optimize a serving system, one important consideration is the relative cost between prefilling and decoding.
Figure~\ref{fig:prefill_dec} shows the relative latency under two model sizes (Yi 34B and Command R+ are open-source GPT3.5 and GPT-4 level, respectively).
We see that as the model becomes large and as the context length becomes longer, the cost gradually shifts from decoding to prefilling.
Also for tasks that require multiple rounds of user interactions (e.g., coding agent and AI companion), their decoding cost is generally higher than tasks of less rounds of conversation (e.g., document QA). 
Given that most of the existing inference optimization such as speculative decoding~\citep{leviathan2023fast, chen2023accelerating} and vLLM~\citep{kwon2023efficient} targets the short-context regime, 
their improvements may \textit{not} be significant enough for long-context deployment. Below we discuss further optimization directions.

\textbf{Optimizing Prefilling}\quad\quad
Since prefilling is compute bound, the space for optimization is not actually very large, and mostly reduce to decreased flops. 
One straightforward flop reduction is local or linear attention (e.g., the sliding-window attention used by Mistral~\citep{jiang2024mixtral}), which removes the quadratic attention term in Eq.~\ref{eq:prefilling}.
However, Fig.~\ref{fig:prefill_dec} shows that for context length of less than 50K, the gain of linear attention is quite limited. Eventually, to reduce the prefilling latency one many still fall back to smaller and shallower models, like the practice of YOCO~\citep{Sun2024YouOC}.

\textbf{Optimizing Decoding}\quad\
Compared to prefilling, the space for optimizing decoding is much larger, not only because there exist a large level of sparsity within the KV cache as we discussed above, but also because of existing techniques like speculative decoding and their long-context variants like TriForce~\citep{sun2024triforce}.
So in general it might be relatively easier to optimize decoding, 
but it is important to keep in mind that for large models of long context, it is the prefilling stage that takes most of the time (see Fig.~\ref{fig:prefill_dec} the Command R+ 200K case).


%% file: tables/tab_existing_work.tex
\begin{table*}[t!]
\centering
\caption{
How existing efforts improves the four performance metrics.
C: concurrency, P: prefilling, D: decoding, S: context switching. 
In general, the layer dimention and the head dimension seem to exhibit high sparsity and might be radically compressed, while the token and the hidden dimension may only be mildly (but still possible) compressed.
}
\begin{tabular}{@{}llll@{}}
\toprule
                    & \bf Desc & \bf Improves&\bf Needle?  \\ 
\midrule  
\bf Layer & & \\ 
CALM~\citep{schuster2022confident} & Early exit based on estimated confidence & C | P | D | S & ? \\ 
CoLT5~\citep{ainslie2023colt5} & Conditonally reducing computation on some layer & C | P | D | S & ? \\ 
LayerSkip~\citep{elhoushi2024layer} & Skipping some layers then verify & C | P | D | S & ?\\ 
YOCO~\citep{Sun2024YouOC} & Use only one global KV cache & C | P | D | S & \checkmark \\ 
\midrule  
\bf Head & & \\ 
\citet{voita-etal-2019-analyzing} & Head pruning based on gating& C | D | S & ? \\ 
GQA~\citep{ainslie2023gqa} & Reusing KV cache for groups of heads & C | D | S & \checkmark\\ 
Retrieval Head~\citep{wu2024retrieval} & Removing non-retrieval heads & C | D | S & \checkmark\\ 
MLA~\citep{DeepSeekAI2024DeepSeekV2AS}  & Using latent head & C | P | D | S  & \checkmark\\ 
\midrule  
\bf Token & & \\ 
H2O~\citep{zhang2024h2o}    & Dropping insignificant tokens after prefilling  & C | D | S  & ? \\
FastGen~\citep{ge2023model}  & Identify important tokens during prefilling & C | D | S    & ? \\
DMC~\citep{nawrot2024dynamic} & Dynamically merge tokens & C | P | D | S & ? \\ 
SnapKV~\citep{li2024snapkv} & Identify important tokens based on user questions & D & \checkmark\\ 
TriForce~\citep{sun2024triforce} & Speculative decoding for long-context & D & \checkmark\\ 
\midrule  
\bf Hidden & & \\ 
KIVI~\citep{liu2024kivi} & KV cache quantization  & C | D | S & ? \\ 
WKVQuant~\citep{yue2024wkvquant} & Weight and KV cache quantization & C | D | S & ? \\ 
\bottomrule
\end{tabular}
\label{tab:existing_work_kv_compression}
\end{table*}

%% file: figures/fig_prefill_dec.tex
\begin{figure}[t!]
    \centering
    \includegraphics[width=\linewidth]{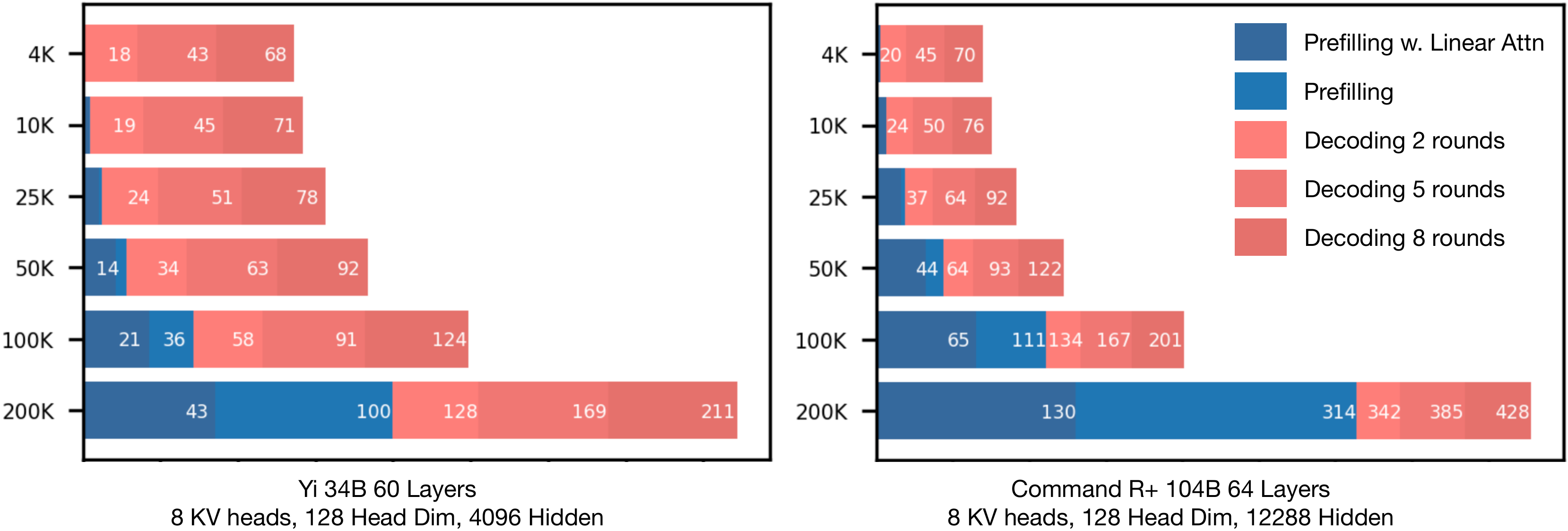}
    \caption{Should one target prefilling or decoding? 
    Here we show how Yi 34B~\citep{young2024yi} (about GPT-3.5 level) and Command R+~\citep{cohere2024command}
    (about GPT-4 level)'s latency of prefilling and decoding (measured in seconds) on different input length and conversation rounds.
    The larger the model, the longer the input context, the better gain from the improved prefilling (e.g., using a 104B model to answer 5 questions of a book); 
    the smaller the model, the more followup questions, the better gain from improved decoding (e.g., talking to a 34B AI companion for 2 hours of 100+ dialog rounds). 
    Also note that when the context is less than 50K, changing the attention from full to linear (as many works count on linear attention to improve efficiency) does \textit{not} significantly improve the overall performance. 
    }
    \label{fig:prefill_dec}
\end{figure}

%% file: 040_conclusion.tex
In this work, we give a detailed analysis of the challenges in deploying long-context transformers. 
Our eventual objective is to reduce the serving cost of 1M context to be as cheap as 4K, such that we can democratize emerging AI applications like video understanding and generative agents. 
We describe a concurrent programming framework to illustrate the end-to-end user interaction session based throughput, and decompose it into four key performance metrics: 
concurrency, prefilling, decoding, and context switching. 
We discuss how common factors influence the four metrics and how existing works focus on different metrics. 
We believe there are great research opportunities to integrate existing efforts to build a strong end-to-end long-context serving system and believe that this work can serve as the foundation for full-stack long-context inference optimization. 